\documentclass[letterpaper, 10 pt, conference]{ieeeconf}  

\IEEEoverridecommandlockouts                              

\usepackage{multicol}
\usepackage{multirow} 
\usepackage{makecell}
\usepackage{graphicx}
\usepackage{amsmath}
\usepackage{colortbl}
\usepackage{amssymb}
\usepackage{booktabs}
\usepackage{caption}
\usepackage{hyperref}
\title{\LARGE \bf
NavDP: Learning Sim-to-Real Navigation Diffusion Policy with Privileged Information Guidance
}

\author{Wenzhe Cai$^{*,1}$, Jiaqi Peng$^{*,1,2}$, Yuqiang Yang$^{1}$, Yujian Zhang$^{3}$, Meng Wei$^{1,4}$ \\ Hanqing Wang$^{1}$, Yilun Chen$^{1}$, Tai Wang$^{1,\dagger}$, Jiangmiao Pang$^{1,\dagger}$
\thanks{$^{1}$Shanghai AI Lab, $^{2}$Tsinghua University, $^{3}$Zhejiang University $^{4}$The University of Hong Kong}\thanks{$^{*}$Equal Contribution, $^{\dagger}$Corresponding Authors}}

\begin{document}

\maketitle
\thispagestyle{empty}
\pagestyle{empty}

\begin{abstract}
Learning to navigate in dynamic and complex open-world environments is a critical yet challenging capability for autonomous robots.
Existing approaches often rely on cascaded modular frameworks, which require extensive hyperparameter tuning or learning from limited real-world demonstration data.
In this paper, we propose Navigation Diffusion Policy (NavDP), an end-to-end network trained solely in simulation that enables zero-shot sim-to-real transfer across diverse environments and robot embodiments. 
The core of NavDP is a unified transformer-based architecture that jointly learns trajectory generation and trajectory evaluation, both conditioned solely on local RGB-D observation.
By learning to predict critic values for contrastive trajectory samples, our proposed approach effectively leverages supervision from privileged information available in simulation, thereby fostering accurate spatial understanding and enabling the distinction between safe and dangerous behaviors. 
To support this, we develop an efficient data generation pipeline in simulation and construct a large-scale dataset encompassing over one million meters of navigation experience across 3,000 scenes. 
Empirical experiments in both simulated and real-world environments demonstrate that NavDP significantly outperforms prior state-of-the-art methods. Furthermore, we identify key factors influencing the  generalization performance of NavDP. The dataset and code are publicly available at
\href{https://wzcai99.github.io/navigation-diffusion-policy.github.io}{\textbf{https://wzcai99.github.io/navigation-diffusion-policy.github.io}}.

\end{abstract}

\begin{figure*}[htp]
    \centering
    \includegraphics[width=0.95\textwidth]{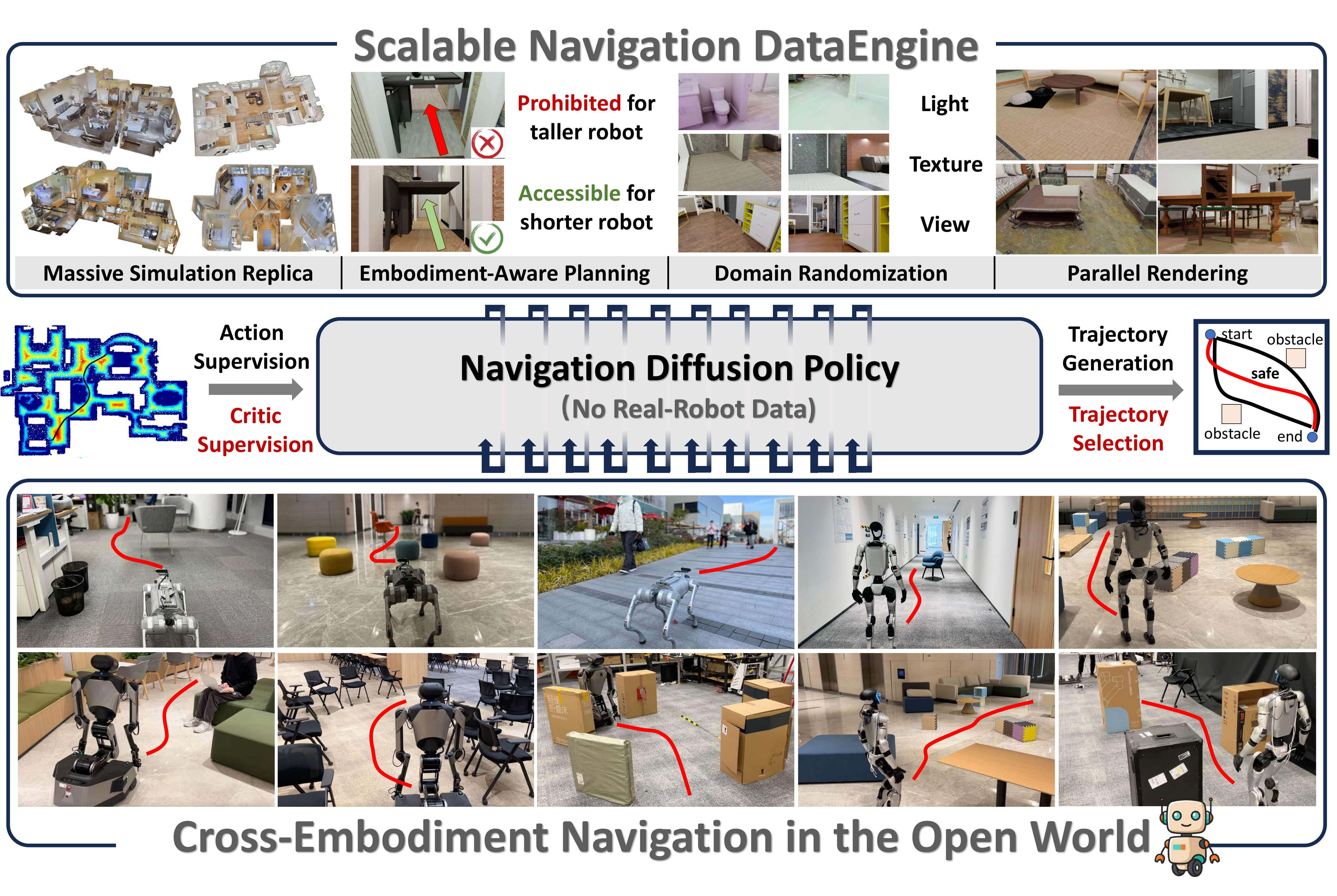}
    \vspace{-0.5cm}
    \captionof{figure}{NavDP is solely trained with simulation data but can achieve zero-shot sim-to-real transfer to different types of robots. By learning from the prioritized knowledge in the simulation data, NavDP forms commonsense spatial understanding ability for navigation task and can adaptively select safe navigation routes towards the goal without any maps.}
    \label{fig:teaser}
    \vspace{-0.5cm}
\end{figure*}
\vspace{-0.2cm}
\section{Introduction}
Navigation in dynamic open world is a fundamental yet challenging skill for robots. For pursuing embodied intelligent generalists, the navigation system is expected to be capable of zero-shot generalizing across different embodiment and unstructured scenes. However, the traditional modular-based methods suffer from system latency and compounding errors which limits their performance, while the scarcity of high-quality data limits the scale-up training and performance of learning-based methods. Although several studies try to address this problem by collecting robot trajectories in the real world~\cite{hirose2019deep,karnan2022socially,hirose2023sacson}, the scaling process is still time-consuming and expensive. 

In contrast, simulation data is diverse and scalable. With large-scale 3D digital replica scenes available~\cite{straub2019replica,chang2017matterport3d,fu20213d,khanna2024habitat,ramakrishnan2021habitat,wang2024grutopia}, we can efficiently generate customized infinite navigation trajectories with different types of observations and goals.
Furthermore, with the increasing diversity of 3D assets and rapid progress of neural rendering algorithms, the long-standing sim-to-real gap problem can also be alleviated. To learn generalized navigation policies, imitation learning ~\cite{meng2025aim,sridhar2024nomad} typically rely on positive demonstrations, but they lack interaction and fail to incorporate negative feedback from the environment. 
In contrast, reinforcement learning (RL)-based methods~\cite{zengpoliformer,eftekhar2024one} can learn from interactions with reward signals, but suffer from low data efficiency.

In this paper, we propose a novel end-to-end transformer-based learning framework to combine the advantages of these two streams, \textbf{Nav}igation \textbf{D}iffusion \textbf{P}olicy \textbf{(NavDP)}, which achieves zero-shot sim-to-real policy transfer and cross-embodiment generalization with only simulation data. 
Our proposed framework leverages the efficiency of imitation learning and the expressiveness of a diffusion process to model the multi-modal distribution of expert demonstrations. To enable counterfactual reasoning, we adapt the concept of the critic value function from reinforcement learning and train NavDP to predict state-action values for both positive and negative trajectories. Our framework can fully take advantages of the privileged information in the simulation from two aspects: On the one hand, the trajectory generation can be trained under the guidance from global-optimal planner within simulation environments. 
On the other hand, the critic function can learn from arbitrary generated contrastive action samples with the global Euclidean Signed Distance Field (ESDF) available in simulation as a fine-grained guidance.
To support large-scale training of NavDP, we developed a highly efficient navigation data engine capable of generating 2,500 trajectories per GPU per day—achieving a 20× improvement in efficiency over real-world data collection. This enables the creation of a comprehensive dataset encompassing over one million meters of robot navigation experience across 3,000 diverse scenes. We conduct extensive empirical evaluations in both simulated and real-world environments. The results demonstrate our proposed NavDP outperforms the previous state-of-the-art approaches by a large margin.

\section{Related Works}
\textbf{Robot Diffusion Policy.} Advanced generative models have shown great potential in capturing multimodal distribution of robot policy learning. The diffusion policy~\cite{chi2023diffusionpolicy} was the first to introduce the diffusion process into manipulation tasks, sparking numerous efforts to enhance its capabilities. These enhancements span various aspects, including state representations~\cite{ke20243d, li2024crossway, wang2024sparse, Ze2024DP3}, inference speed~\cite{prasad2024consistency, wang2024one}, and deployment across diverse robot applications~\cite{huang2024diffuseloco, zhang2023generative, xu2023xskill, wang2024dexcap}. However, as diffusion policies operate within an offline imitation learning framework, achieving strong real-world performance often depends on real-world teleoperation datasets, which are labor-intensive and challenging to scale up. In contrast, our approach develops robot policies entirely from scalable simulation datasets. To enhance generalization and ensure safety during sim-to-real transfer, we introduce a critic function to estimate the safety of policy outputs. This mechanism leverages prioritized simulation data to enable the diffusion policy to understand the consequences of actions, improving both safety and performance.

\textbf{End-to-End Visual Navigation Models.} Recent end-to-end visual navigation models have demonstrated significant potential in cross-embodiment adaptation and multi-task generalization~\cite{shah2023gnm, shah2023vint, yang2023iplanner, roth2024viplanner, cai2024bridging, ehsani2024spoc, zhang2024navid, zhang2024uni}. These approaches tackle navigation challenges at various levels of abstraction. Vision-Language-Action (VLA) models~\cite{ehsani2024spoc, zhang2024navid, zhang2024uni, cheng2024navila} offer flexibility by leveraging language instructions for task specification. In contrast, end-to-end local navigation models excel in cross-embodiment generalization and demonstrate superior adaptability with real-time inference in open-world environments~\cite{shah2023gnm, shah2023vint, sridhar2024nomad}. In this paper, we focus on developing efficient end-to-end cross-embodiment navigation system-1 which can seamlessly attach to the VLM for generalizable navigation task execution skill in dynamic open-world.
\section{DataEngine}
\textbf{Robot Model.} We build the robot as a cylindrical rigid body with a two-wheel differential drive model for cross-embodiment generalizability. The navigation safe radius of the robots is set to $r_{b}=0.25\text{m}$. 
To imitate the variation of observation views across different robot embodiments, we assume one RGB-D camera is installed on the top of the robot and the height of the robot $h_{b}$ is randomized in the range $(0.25\text{m}, 1.25\text{m})$. Objects that are higher than the camera's configuration height are not considered as obstacles during the navigation trajectory planning process.
To ensure the local navigable area remains visible within the field of view, the camera's pitch angle is randomized within the range $(-30^{\circ},0^{\circ})$, depending on the robot's height. We use two configurations to set the camera field of views, one follows RealSense D435i with horizontal field of view (HFOV) and vertical field of view (VFOV) set to $(69^{\circ},42^{\circ})$, and the other follows Zed 2 with FOV set to $(110^{\circ},70^{\circ})$. 

\textbf{Trajectory Generation.}
To generate collision-free robot navigation trajectories, we first convert the scene meshes into a voxel map with a voxel size of $0.05\text{m}$ to estimate the Euclidean Signed Distance Field (ESDF) of the navigable areas. Navigable areas are defined as voxel elements with $z$-axis coordinates below the threshold $h_\text{nav}$, while obstacle areas are defined as voxel elements with $z$-axis coordinates exceeding the threshold $h_\text{obs}$. The thresholds $h_\text{nav}$ and $h_\text{obs}$ vary across scenes and depend on the robot height $h_b$. 
Voxels with distance values lower than the robot radius $r_b$ are truncated to prevent collisions.
The ESDF map of the navigable area is downsampled to $0.2\text{m}$ resolution to facilitate efficient A* path planning. Navigation start and target points are selected randomly on the navigable area, and the A* algorithm generates a planned path $\tau^* = [(x_0, y_0), (x_1, y_1), (x_2, y_2), \dots, (x_k, y_k)]$. For each waypoint $(x_n, y_n)$, a greedy search is performed in a local area of the original ESDF map to refine the position by maximizing the distance to nearby obstacles. This refinement process shifts waypoints further from obstacles. Finally, the refined waypoints are smoothed into a continuous navigation trajectory using cubic spline interpolation.

\textbf{Scene Assets and Render Engine.} Following the pipeline described in the previous section, we can generate a large-scale dataset of robot navigation trajectories and corresponding RGB-D rendering results across diverse scenes. We use BlenderProc~\cite{denninger2020blenderproc} to render photorealistic RGB and depth images along the navigation trajectories. We collect navigation trajectories from over 3,000 scenes selected from 3D-Front~\cite{fu20213d}, HSSD~\cite{khanna2024habitat}, HM3D~\cite{ramakrishnan2021habitat}, Replica~\cite{straub2019replica}, Gibson~\cite{xiazamirhe2018gibsonenv}, and Matterport3D~\cite{chang2017matterport3d}. For each scene, we sample 100 pairs of starting points and destinations. We adapt several domain randomization techniques to further improve the data diversity, which contains light condition randomization, view randomization as well as texture randomization.
After data filtering, the final dataset comprise over 200K trajectories covering more than 1M meters. Compared with the previous navigation dataset, our data dominates in diversity and collection efficiency as shown in Table~\ref{tab:dataset}. The dataset will be open-sourced in the near future.
\vspace{-0.2cm}
\begin{table}[ht!]
    \renewcommand\arraystretch{1.5}
    \centering
    \fontsize{7}{6}\selectfont
    \begin{tabular}{cccccc}
    \toprule
    \textbf{Dataset} & \textbf{Scene} & \textbf{Distance (Km)} & \textbf{Hour} & \textbf{Image} & \textbf{Collection} \\   
    \midrule
    GoStanford~\cite{hirose2019deep} & 27 & 25.5 & 16.7 & 178K & Teleop \\
    \midrule
    RECON~\cite{shah2021rapid} & 9 & 152.5 & 40 & 610K & Autonomous \\
    \midrule
    SCAND~\cite{karnan2022socially} & 1 & 40 & 8.7 & 100K & Teleop \\
    \midrule
    SACSoN~\cite{hirose2023sacson} & 5 & 58 & 75 & 241K & Autonomous \\
    \midrule
    AMR~\cite{meng2025aim} & 54 & - & - & 7.5M & Simulation \\
    \midrule
    \textbf{NavDP (ours)} & \textbf{3154} & \textbf{1627.1} & \textbf{452} & \textbf{40M} & Simulation \\
    \bottomrule
    \end{tabular}
    \caption{Quantitative comparison of navigation datasets.}
    \label{tab:dataset}
    \vspace{-0.5cm}
\end{table}

\section{Navigation Diffusion Policy}
\begin{figure}[htp!]
    \centering
    \includegraphics[width=0.5\textwidth]{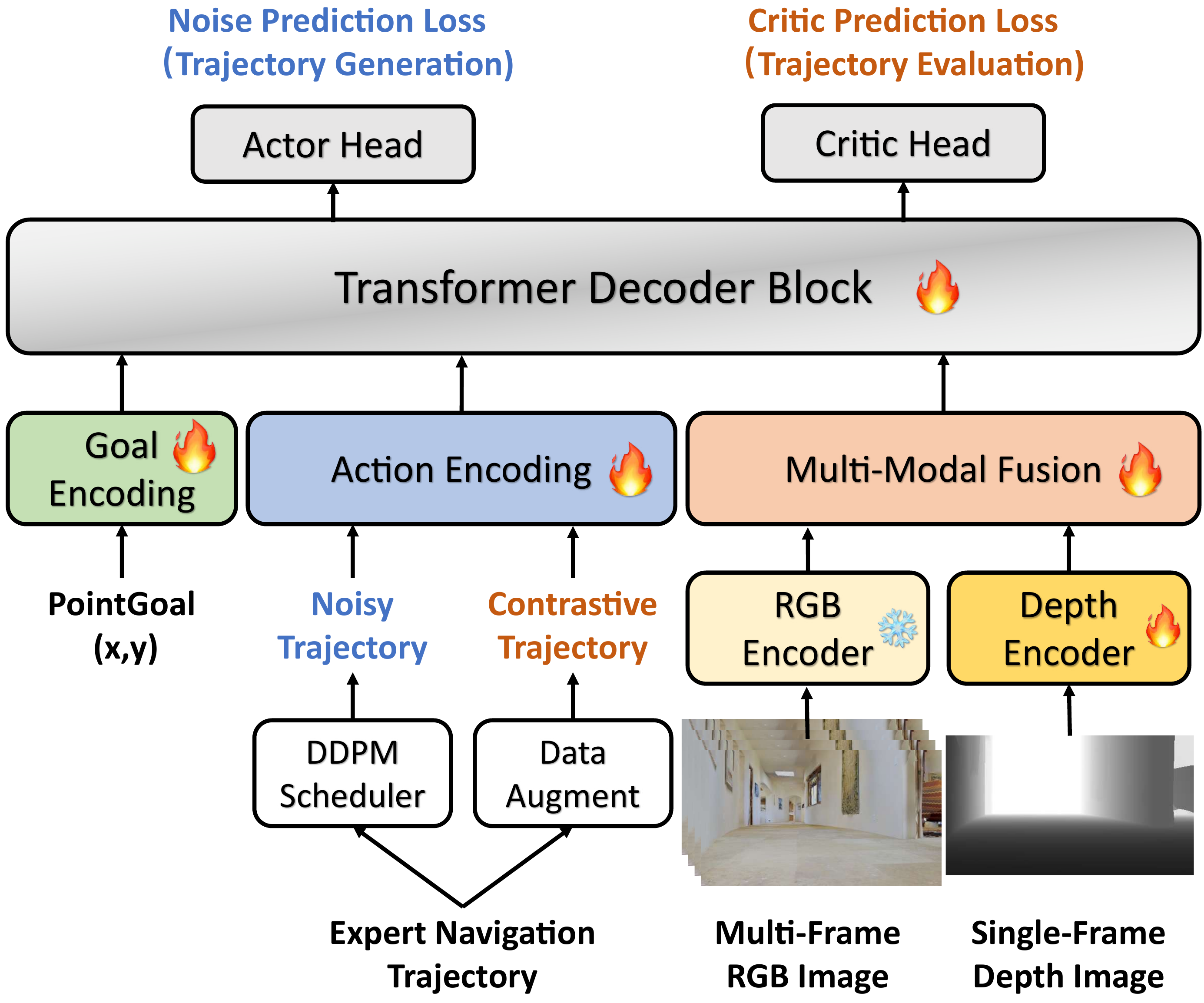}
    \vspace{-0.5cm}
    \captionof{figure}{Overview of the network architecture. NavDP is conditioned on RGB-D observations and navigation trajectories. During training, Gaussian noise is added to the ground-truth trajectory according to the DDPM scheduler, and the actor head is trained to predict the injected noise. Simultaneously, the ground-truth trajectories are augmented to create both collision-free and collision scenarios, and the critic head is trained to assign contrastive scores to these trajectories.}
    \label{fig:navdp-network}
    \vspace{-0.2cm}
\end{figure}

NavDP consists of a multi-modal encoder to fuse RGB and depth observations and a unified transformer-based network for both trajectory generation and critic value prediction. The NavDP network architecture is shown in Figure~\ref{fig:navdp-network}. The trajectory generation aims to plan $M$ dense waypoints for robots to follow while the critic value prediction aims to predict 
scores corresponding to the safety of trajectories.

\textbf{Multi-Modal Encoder.} NavDP processes RGB-D images and a navigation goal as inputs. To incorporate historical information, we feed multi-frame RGB images of length $N$ and use a pre-trained DepthAnything~\cite{yang2024depth} encoder to extract 256 patch tokens from each RGB frame. To ensure alignment with absolute physical scale for trajectory generation, we introduce an additional ViT encoder—trained from scratch—to process a single frame of depth observation, which also produces 256 tokens. As the depth input may suffer from sim-to-real gap, here we only use the depth within range $(0.1\text{m},5\text{m})$. To fuse the RGB-D inputs, we apply lightweight transformer decoder layers with learnable queries, compressing the original $(N+1)\times 256$ tokens into $N\times 16$ compact tokens.
The navigation goal follows the PointGoal task definition, where a 2-dimensional vector 
$(x_{g},y_{g})$ represents the goal's relative coordinates with respect to the current state. 
We employ MLP layers to encode the navigation goal and project it into the same dimensional space as the RGB-D tokens for subsequence process. For no-goal task, we use full-zero tensor as the goal embedding.

\begin{figure*}[htp!]
    \centering
    \includegraphics[width=0.95\textwidth]{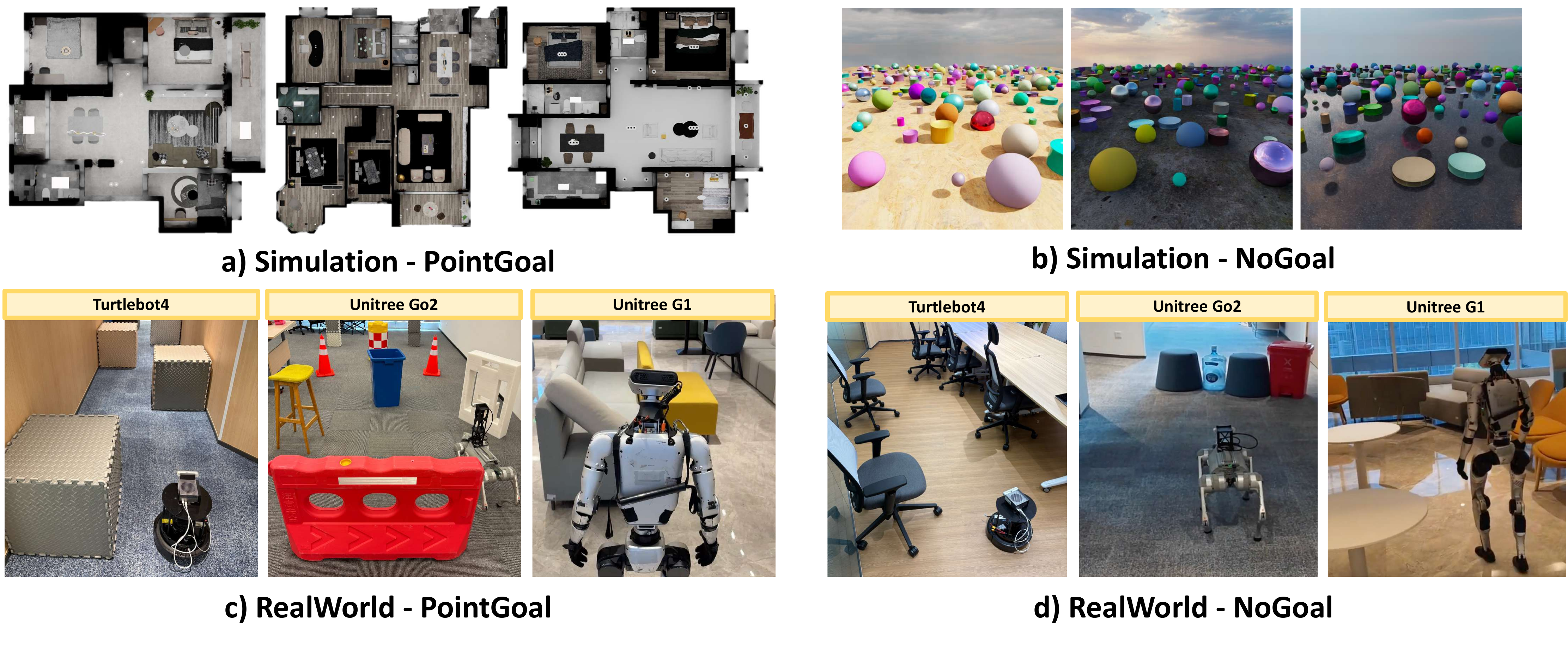}
    \vspace{-0.3cm}
    \captionof{figure}{An overview of the evaluation scenes, including both simulation and real-world. 
    In simulation, we adapt 10 home scenes, 10 commercial scenes for point-goal task, and 10 scenes with clutter layout for no-goal task. In real-world, we evaluate different navigation policy on Turtlebot4, Unitree Go2, G1 and Galaxea R1 in both indoor and outdoor scenes.}
    \vspace{-0.4cm}
    \label{fig:navdp-eval}
\end{figure*}

\textbf{Unified Policy Transformer.} We develop a simple yet effective transformer decoder-based architecture that supports both diffusion-based trajectory generation and trajectory evaluation.
For trajectory generation, the objective is to predict the injected noise conditioned on a noisy trajectory. For trajectory evaluation, the goal is to predict a score conditioned on an arbitrary trajectory.
To this end, we use an MLP-based action encoder to extract trajectory embeddings, which serve as queries in the cross-attention mechanism. The fused RGB-D tokens, along with a token representing the diffusion timestep, act as keys and values in the attention process.
We employ multiple transformer decoder layers to process the input tokens and use two separate output heads for the two tasks. All network weights are shared between tasks; the distinction lies in the input queries and the attention masks applied to the keys and values.
In the trajectory generation task, queries are extracted from noisy trajectories with noise added according to the DDPM~\cite{ho2020denoising} scheduler, and cross-attention attends to all keys and values. In the trajectory evaluation task, queries are derived from expert-demonstrated trajectories with random rotation augmentations, and the cross-attention excludes the timestep token.
During inference, NavDP first generates a batch of candidate trajectories using the trajectory generation head and then selects the best trajectory via the evaluation head, thereby enabling safer task execution.

\textbf{Training Details.} The training objectives for trajectory generation task is weighted sum of mean squared error (MSE) loss of both predicted noises for point-goal and no-goal tasks. Denote $k$ as the denoising steps, $\epsilon_{k}$ as the sampled noise following DDPM scheduler at timestep $k$, $d_{t}$ as the current depth image, $I_{t-N:t}$ as the RGB observations, $g_{t}$ as the navigation goal, $\tau$ as the expert demonstration trajectories without noise, $\mathcal{D}$ as the entire dataset, the loss function for actor head can be written as follows:
\begin{equation}
    \mathcal{L}_{\text{act}}^{\text{ng}} = \mathbb{E}_{\tau,g,d,I\sim \mathcal{D}}[(\epsilon_{k} - \epsilon_{\theta}(\tau+\epsilon_{k+1}, k, d_{t}, I_{t-N:t}))^{2}]
\end{equation}
\begin{equation}
    \mathcal{L}_{\text{act}}^{\text{pg}} = \mathbb{E}_{\tau,g,d,I\sim \mathcal{D}}[(\epsilon_{k} - \epsilon_{\theta}(\tau+\epsilon_{k+1}, k, g_{t}, d_{t}, I_{t-N:t}))^{2}]
\end{equation}
\begin{equation}
    \mathcal{L}_{\text{act}} = \alpha \cdot \mathcal{L}_{\text{act}}^{\text{ng}} + \beta \cdot  \mathcal{L}_{\text{act}}^{\text{pg}}
\end{equation}

We set $\alpha=0.5$ and $\beta=0.5$ by default. $\epsilon_{\theta}$ is the entire noise prediction network.
For the trajectory evaluation task, we define the label critic value with respect to both the absolute ESDF value and difference of ESDF value along the trajectory waypoints. Concretely, denote the augmented expert demonstration trajectory as $\hat{\tau}$, the ESDF value at $m$-th waypoint on the augmented trajectory as $d_{\hat{\tau}}^{m}$, the label critic value is defined as follows:
\begin{equation}
    V(\hat{\tau}) = \gamma \cdot \sum_{m=0}^{M}(d_{\hat{\tau}}^{m+1} - d_{\hat{\tau}}^{m})+ \lambda \cdot \frac{1}{M}\sum_{m=0}^{M}\mathbb{I}(d_{\hat{\tau}}^{m} < d_{safe})
\end{equation}
$d_{safe}$ is a threshold representing the collision radius.
Then, denote the score prediction network as $V_{\theta}$, the loss function for critic head can be written as the follows:
\begin{equation}
    \mathcal{L}_{\text{critic}} = \mathbb{E}_{I,d,\tau \sim \mathcal{D}}[V(\hat{\tau}) - V_{\theta}(I_{t-N:t},D_{t},\hat{\tau})]
\end{equation}
And the whole NavDP network is jointly trained in one stage with respect to the sum over $\mathcal{L}_{\text{act}}$ and $\mathcal{L}_{\text{critic}}$. Some other hyper-parameters are shown in Table~\ref{tab:parameter}.
\begin{table}[ht!]
\vspace{-0.1cm}
    \renewcommand\arraystretch{1.5}
    \centering
    \fontsize{9}{8}\selectfont
    \begin{tabular}{cc}
        \toprule
         \multirow{1}{*}{\textbf{Hyper-parameters}}& \multicolumn{1}{c}{\textbf{Value}} \\
        \cmidrule(r){1-2}
      Diffusion Step & 10 \\
      Prediction Waypoints $M$ & 24 \\
      RGB History Size $N$ & 8 \\
      Safe Distance Threshold $d_{safe}$ & 0.5 \\
      Training GPU & 32 A100 cards \\ 
      Training Batchsize & 2048 \\
      Learning Rate & 1e-4 \\
      Learning Rate Decay & Linear \\
      Training GPU Hours & 24$\times$32 \\
      \bottomrule
    \end{tabular}
    \label{tab:training-details}
    \caption{Table of hyper-parameters of training.}
    \vspace{-0.4cm}
    \label{tab:parameter}
\end{table}

\begin{table*}[ht!]
    \renewcommand\arraystretch{1.4}
    \centering
    \fontsize{10}{9}\selectfont
    \begin{tabular}{ccc|cccc}
    \toprule
    \multirow{2}{*}{\textbf{Point-Goal Nav}} & \multicolumn{2}{c}{\textbf{Dingo-Sim}} & 
    \multicolumn{4}{c}{\textbf{Cross-Embodiment Real-world}} \\ 
    \cmidrule(r){2-7}
     & SR($\uparrow$) & SPL($\uparrow$) & SR($\uparrow$) & Turtlebot($\uparrow$) & Unitree-Go2($\uparrow$) & Unitree-G1($\uparrow$) \\
     \midrule
     DD-PPO~\cite{wijmansdd} & 8.6 & 8.5 & - & - & - & -  \\
     \midrule
     EgoPlanner~\cite{zhou2020ego} & - & - & 40.0 & 5/10 & 3/10 & - \\
     \midrule
     iPlanner~\cite{yang2023iplanner} & 54.1 & 51.2 & 16.7 & 0/10 & 5/10 & 0/10\\
     \midrule
     ViPlanner~\cite{roth2024viplanner} & 60.9 & 58.6 &  53.3 & 5/10 & 4/10 & \textbf{7/10}\\
     \midrule
     \textbf{NavDP (Ours)} & \textbf{67.2} & \textbf{62.6} &  \textbf{76.7} & \textbf{9/10} & \textbf{7/10} & \textbf{7/10} \\
     \bottomrule
    \end{tabular}
    \caption{Quantitative Evaluation Results of Point-Goal Navigation. We report both detailed success rate for each embodiment and the overall success rate in real-world experiments. Our proposed NavDP outperforms the previous state-of-the-art approach in simulation by \textbf{6.3\%} success rate and \textbf{4.0\%} SPL, \textbf{23.4\%} success rate in real-world evaluation.}
    \label{tab:experiment-pointgoal}
\end{table*}

\begin{table*}[ht!]
    \renewcommand\arraystretch{1.4}
    \centering
    \fontsize{10}{9}\selectfont
    \begin{tabular}{ccc|cccc}
    \toprule
    \multirow{2}{*}{\textbf{No-Goal Nav}} & \multicolumn{2}{c}{\textbf{Dingo-Sim}} & \multicolumn{4}{c}{\textbf{Cross-Embodiment Real-world}} \\
    
    \cmidrule(r){2-7}
     & Time($\uparrow$) & Area($\uparrow$) & Time($\uparrow$) & Turtlebot($\uparrow$) & Unitree-Go2($\uparrow$) & Unitree-G1($\uparrow$) \\
     \midrule
     GNM~\cite{shah2023gnm} & 12.5 & 29.6 & 15.2 & 9.9 & 12.9 & 23.0 \\
     \midrule
     ViNT~\cite{shah2023vint} & 18.9 & 46.6 & 13.1 & 15.6 & 15.8 & 8.0 \\
     \midrule
     NoMad~\cite{sridhar2024nomad} & 36.6 & 85.7 & 29.3 & 17.4 & 37.2 & 33.5 \\
     \midrule
     \textbf{NavDP (Ours)} & \textbf{106.2} & \textbf{274.1} & \textbf{112.9} & \textbf{114.2} & \textbf{143.3} & \textbf{81.3} \\
     \bottomrule
    \end{tabular}
    \caption{Quantitative Evaluation Results of No-Goal Navigation. We report both detailed exploration time for each embodiment and the overall time. Our proposed NavDP achieves a nearly \textbf{2.9x} performance in exploration time and \textbf{3.1x} in exploration area better than the previous method in simulation, and \textbf{3.8x} performance in real-world exploration time.}
    \vspace{-0.5cm}
    \label{tab:experiment-nogoal}
\end{table*}

\section{Experiments}
\subsection{Evaluation and Metrics}
We evaluate our approach with point-goal navigation and no-goal exploration tasks across 4 different robots in both simulation and real-world. 
We build the simulation benchmark based on IsaacSim which offers high-fiedlity physical simulation and use a wheeled robot - ClearPath Dingo as the navigator.
For point-goal navigation task, we collect 20 realistic scenes (10 home, 10 commercial) from GRUtopia~\cite{wang2024grutopia}, covering a wide range of scenarios including home, hospital, supermarket, etc. 
For no-goal navigation task, we generate 10 challenging cluttered scenes with random obstacles as the for evaluation. We evaluate 2,000 episodes for point-goal task and 1,000 episodes for no-goal task.

In real-world benchmark, we evaluate the performance of Unitree Go2, Turtlebot4 and Unitree G1.
For the point-goal navigation task, we setup 3 different indoor scenarios with challenging layout and evaluate 10 episodes for each scene with one embodiment.
For the no-goal exploration task, we evaluate the cross-embodiment generalization in another 3 large indoor scenarios, including corridor, hall and meeting room.
A brief visualization of the simulation and real-world evaluation scenarios are provided in Figure~\ref{fig:navdp-eval}. In point-goal navigation task, we evaluate two metrics: \textbf{Success Rate (SR)} and \textbf{Success Weighted by Path Length (SPL)}, which measures task completion ratio and path efficiency. 
In no-goal navigation task, we evaluate another two metrics - \textbf{Time} and \textbf{Area}, which represents the
the average time in seconds before one collision happens and the average exploration areas. Both help evaluates the overall collision avoidance skill and planning consistency.

\subsection{Experiment Analysis}
In this section, we aim to address the following research questions through both quantitative and qualitative experimental results:
\begin{itemize}
    \item \textbf{Q1}: How well does the proposed NavDP generalize across different robot platforms?
    \item \textbf{Q2}: What are the advantages of our method compared to baseline approaches?
    \item \textbf{Q3}: How well does our method generalize to in-the-wild indoor and outdoor environments?
    \item \textbf{Q4}: What are the key factors that influence the overall performance of the model?
    \item \textbf{Q5}: Is the domain randomization in navigation data essential for achieving cross-embodiment generalization?
\end{itemize}

\begin{figure*}[ht!]
    \centering
    \includegraphics[width=0.90\textwidth]{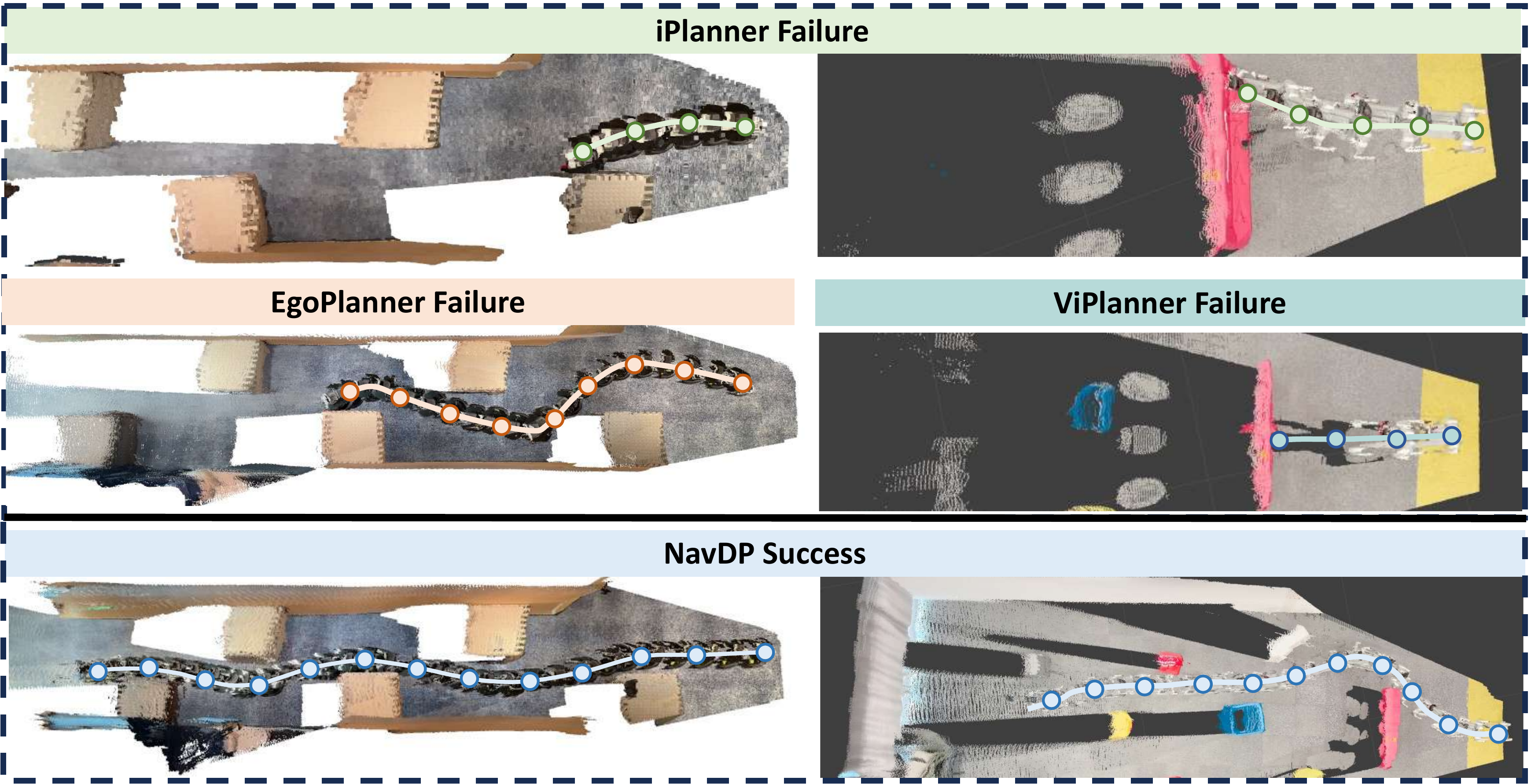}
    \vspace{-0.1cm}
    \captionof{figure}{Visualization of comparison among navigation approaches. Two common failure mode of baselines are displayed: (1) Collision caused by short-horizon memory and planning inconsistency. (2) Faked by the irregular object geometry. Our proposed NavDP can robustly handle both scenarios. The visualization is achieved by $Pi^{3}$~\cite{wang2025pi3}.}
    \vspace{-0.3cm}
    \label{fig:navdp-failure}
\end{figure*}

For \textbf{Q1} and \textbf{Q2}, we compare our proposed NavDP with a range of baseline methods across both navigation tasks. The baselines include learning-based approaches — GNM~\cite{shah2023gnm}, ViNT~\cite{shah2023vint}, NoMaD~\cite{sridhar2024nomad}, DD-PPO~\cite{wijmansdd}, iPlanner~\cite{yang2023iplanner}, and ViPlanner~\cite{roth2024viplanner}, as well as the planning-based approach EgoPlanner~\cite{zhou2020ego}.
In the PointGoal navigation task, NavDP outperforms the previous state-of-the-art learning-based method by \textbf{6.3\%} in Success Rate (SR) within simulation and achieves an average improvement of \textbf{23.0\%} in cross-embodiment real-world experiments (as shown in Table~\ref{tab:experiment-pointgoal}). In contrast, despite the baseline method DD-PPO has been extensively trained with reinforcement learning in the Habitat simulator, it demonstrates poor generalization to out-of-distribution scenarios with different action spaces and camera configurations. Compared with other approaches, our proposed NavDP offers three key advantages: 
\begin{itemize}
    \item \textbf{Temporal Consistency}: iPlanner and ViPlanner rely on single-frame visual input, which restricts temporal consistency in trajectory planning. This limitation often leads to task failure or inefficient behaviors, particularly in the Turtlebot experiment, where the robot camera has already passed obstacles but the body is not. Then, a severe change in path planning can lead to collision.
    \item \textbf{Robustness to Depth Noise}: Traditional planning-based approaches are highly sensitive to noisy depth sensing. In such cases, global mapping errors can result in overly conservative plans, while short-horizon local maps can also cause planning inconsistencies and collisions.
    \item \textbf{Resistance to Depth Illusions}: As depth-centric methods, both iPlanner and ViPlanner are susceptible to misleading object geometries. In the Unitree-Go2 experiments, we place obstacles with holes in front of the robot. Both methods failed to interpret the geometry correctly and trying to walk through the obstacles.
\end{itemize}

Quanlitive visualization of the results from different approaches for point-goal navigation task is presented in Figure~\ref{fig:navdp-failure}. The detail navigation process is provided in the supplimentary video. 
Besides, our proposed method also surpass the baseline methods in no-goal navigation task by a large margin: In simulation, NavDP achieves \textbf{2.9x} performance in average exploration time and \textbf{3.1x} performance in average exploration area than NoMad in simulation and \textbf{3.8x} exploration time in real-world, demonstrating a strong zero-shot generalization to out-of-distribution scenarios. 

For \textbf{Q3}, we deploy NavDP on the Unitree Go2, Unitree G1 and Galaxea R1 robot and visualize the top-2 generated trajectories with the best critic values as shown in Figure~\ref{fig:navdp-visualizaton}. Although the observation views, the existance of pedestrain interference, camera field of views, varying light conditions, the existence of motion blur dramatically make the observation images different from the training dataset, our proposed method still generalize well and can achieve long-horizon navigation without any collision and human intervention over 100 meters. More examples are shown in the appendix video.

For \textbf{Q4}, we conduct a comprehensive ablation study on the point-goal navigation benchmark to investigate three critical factors that may influence the overall performance of NavDP: (1) input modalities, (2) the function of the critic prediction, and (3) the choice of training objectives. Specifically, we evaluate six variants of the original NavDP model:

\begin{table}[ht!]
    \renewcommand\arraystretch{1.2}
    \centering
    \fontsize{9}{9}\selectfont
    \begin{tabular}{ccccc}
    \toprule
    \multirow{2}{*}{\textbf{PointNav}} & \multicolumn{2}{c}{\textbf{Sim-Home}} & \multicolumn{2}{c}{\textbf{Sim-Commercial}}\\ 
    \cmidrule(r){2-5}
     & Success & SPL & Success & SPL \\
     \midrule
     w/o Depth$^{*}$ & 47.8 & 44.3 & 66.1 & 63.7 \\
     \midrule
     w/o RGB$^{*}$ & 53.9 & 49.6 & 70.3 & 66.7  \\
     \midrule
     w/o Multiframe RGB$^{*}$ & 56.9 & 51.7 & 72.0 & 68.2 \\
     \midrule
     w/o Selection & 53.1 & 49.0 & 65.6 & 62.5 \\
     \midrule
     w/o Augmentation$^{*}$ & 57.3 & 52.2 & 73.4 & 69.2 \\
     \midrule
     w/o No-Goal$^{*}$ & 56.8 & 51.6 & 73.5 & 69.9 \\
    \midrule
    \textbf{Original NavDP} & \textbf{60.3} & \textbf{54.7} & \textbf{74.1} & \textbf{70.5} \\
     \bottomrule
    \end{tabular}
    \caption{Quantitative results of ablation experiments.}
    \vspace{-0.5cm}
    \label{tab:ablation}
\end{table}

\begin{itemize}
\item \textbf{w/o Depth$^{*}$}: An RGB-only version of the NavDP model, with the depth input branch removed.
\item \textbf{w/o RGB$^{*}$}: A depth-only version of the NavDP model, with the RGB input branch removed.
\item \textbf{w/o Multiframe RGB$^{*}$}: A single-frame RGB-D variant of NavDP, replacing multi-frame RGB input with a single-frame representation.
\item \textbf{w/o Selection}: A variant of NavDP where the trajectory selection is randomized instead of using the critic-based selection scheme.
\item \textbf{w/o Augmentation$^{*}$}: A version of NavDP where collision trajectories are excluded during critic head training, disabling the trajectory augmentation scheme.
\item \textbf{w/o No-Goal$^{*}$}: A variant of NavDP trained without the no-goal-based auxiliary trajectory prediction objectives.
\end{itemize}

Variants marked with an asterisk ($^*$) indicate that the model is retrained using the same configuration and dataset as the original NavDP. In contrast, variants without an asterisk share the same model weights as the original NavDP. The ablation results are presented in Table~\ref{tab:ablation}. These results highlight three key technical conclusions:

\begin{figure*}[ht!]
    \centering
    \includegraphics[width=1.0\textwidth]{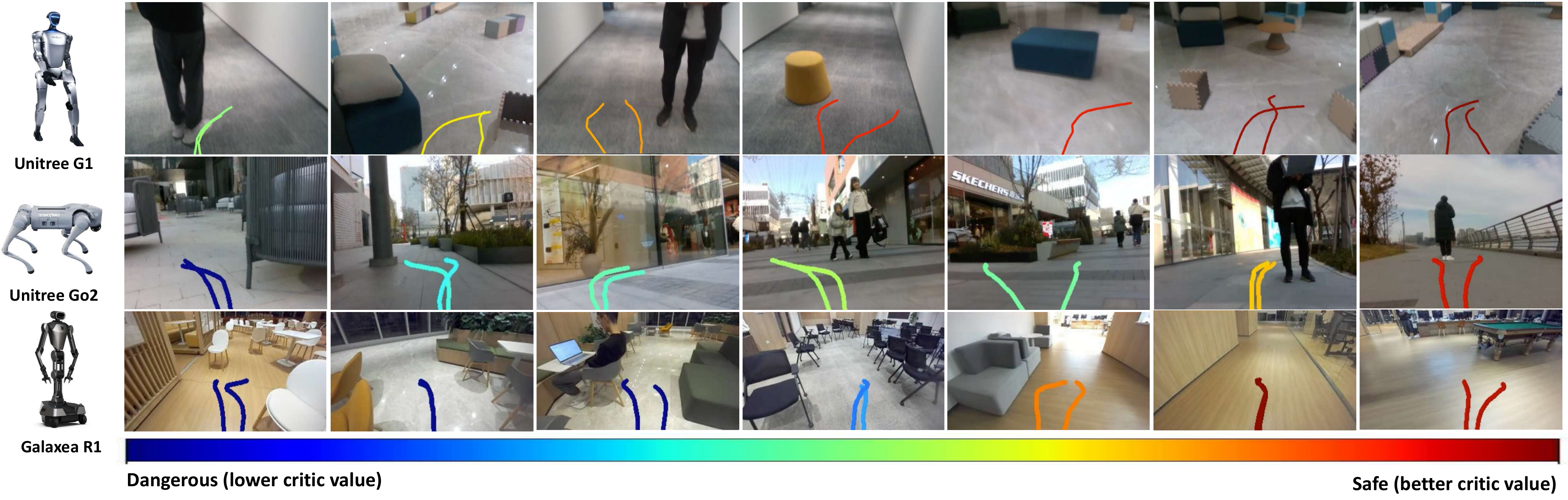}
    \vspace{-0.5cm}
    \captionof{figure}{Trajectory visualization of on different robots. We project the predicted trajectories back to the image space and colorize them according to the corresponding critic values. The \textcolor{blue}{bluer} trajectories indicate higher risk, whereas the \textcolor{red}{redder} trajectories represent safer paths.}
    \vspace{-0.4cm}
    \label{fig:navdp-visualizaton}
\end{figure*}

\begin{itemize}
    \item RGB-D fusion is essential for a more robust navigation performance. Without the depth as input, the overall performance drops \textbf{10.3\%} in success rate and without the rgb images as input, the overall performance drops \textbf{5.1\%}. Multi-frame of RGB input also contributes \textbf{2.8\%} improvements in success rate.
    \item Critic function plays an important role for improving the planning safety. With the same model weights but removing the critic-based trajectory selection, the success rate decreases \textbf{7.8\%} in simulation.  Learning critic function from contrastive samples is also important. After removing the trajectory augmentation for critic training, the success drops \textbf{3.0\%} in home scenes.
    \item The no-goal task is a useful auxiliary task that can be jointly trained. With the no-goal task training objectives, the overall point-goal navigation performance increase \textbf{2.1\%} in success rate and \textbf{1.8\%} in SPL.
\end{itemize}

\begin{figure}[ht!]
    \centering
    \includegraphics[width=0.48\textwidth]{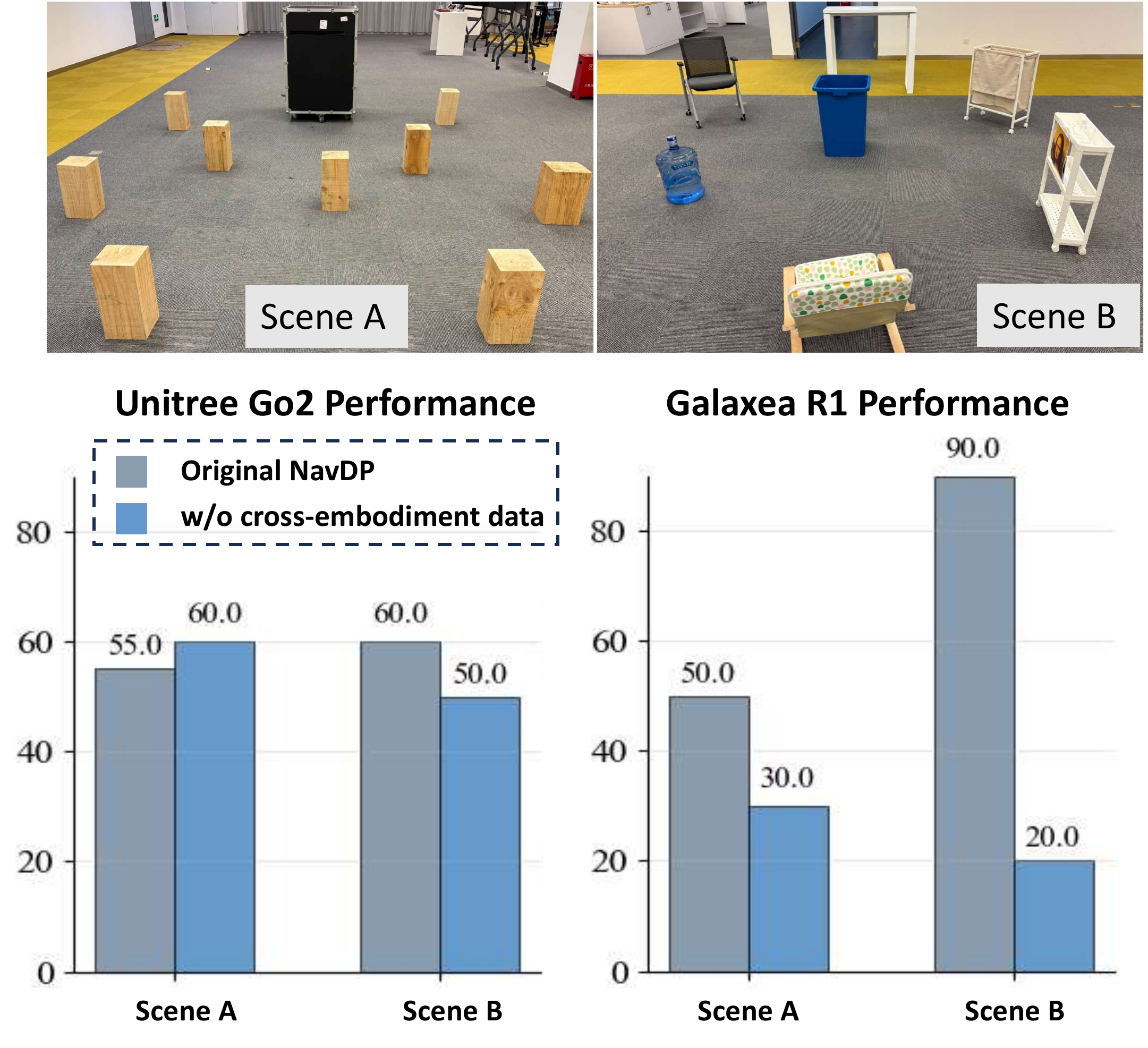}
    \vspace{-0.5cm}
    \captionof{figure}{Ablation study result on cross-embodiment dataset. The top row demonstrates the evaluation scenes while the bottom shows the performance. Without the cross-embodiment data, the performance on Galaxea R1 drops \textbf{70\%} in success rate. }
    \vspace{-0.6cm}
    \label{fig:ablation-cross}
\end{figure}

For Q5, we train a variant of NavDP model only with the data collected at low robot heights $(<0.5m)$ and camera view to prove whether the domain randomization contributes to the cross-embodiment generalization. Then, we setup two challenging scenarios with cluttered layout and evaluates point-goal navigation task performance of two models with both Unitree-Go2 and Galaxea-R1 robots. Each scene is evaluate for 20 episodes for each embodiment. As two robots are different in heights, their path planning strategy should be different: For the Unitree-Go2, it would be an efficient way to walk under the table in SceneB, but the Galaxea R1 must take a detour to avoid the table. The quantitative results are shown in Figure~\ref{fig:ablation-cross}.
We find that without the cross-embodiment data, the detour skill for the table is not exhibited, thus leading to a success rate drops from \textbf{90\%} to \textbf{20\%}, while the overall performance on Unitree-Go2 is maintained.

\section{Conclusion \& Future Works}
In this paper, we introduce a novel navigation diffusion policy (NavDP) that achieves strong zero-shot sim-to-real and cross-embodiment generalization performance. Our policy demonstrates real-time path-planning and collision-avoidance abilities under both static and dynamic scenarios. Two key ingredients contributes to the NavDP performance. The first is our proposed automatic data generation pipeline and large-scale simulation navigation dataset. The second is the design of the entire network with efficient RGB-D fusion and contrastive critic training. Our NavDP provides a novel perspective in building versatile end-to-end navigation policies and a strong backbone for further improvements.

For future work, we plan to explore efficient post-training strategies to further enhance performance—an essential step for real-world deployment. Additionally, we aim to extend NavDP to support a wider range of navigation goals, particularly those expressed through natural language instructions. Finally, integrating a global memory mechanism to enable long-term exploration and more holistic navigation behavior represents another promising direction for future research.




\clearpage
\bibliographystyle{IEEEtran}
\bibliography{root}

\end{document}